\documentclass[letterpaper, 10 pt, conference]{ieeeconf} 
\usepackage{amsmath,amsfonts}
\usepackage{algorithmic}
\usepackage{algorithm}
\usepackage{array}
\usepackage[caption=false,font=normalsize,labelfont=sf,textfont=sf]{subfig}
\usepackage{textcomp}
\usepackage{stfloats}
\usepackage{url}
\usepackage{verbatim}
\usepackage{graphicx}
\usepackage{cite}
\usepackage{nomencl}
\usepackage{ntheorem}   
\usepackage{lipsum}     
\usepackage{graphicx}

\usepackage{xcolor}
\usepackage{hyperref}

\hypersetup{
	colorlinks=true,
	linkcolor=blue,
	filecolor=green, 
	urlcolor=black,
	citecolor=orange
}
\usepackage{amssymb}
\usepackage{bm}

\DeclareRobustCommand{\uvec}[1]{{%
		\ifcsname uvec#1\endcsname
		\csname uvec#1\endcsname
		\else
		\bm{\mathbf{#1}}%
		\fi
}}

\IEEEoverridecommandlockouts                              

\usepackage{graphicx} 
\usepackage{amsmath} 

\title{\LARGE \bf
Developing a Mono-Actuated Compliant GeoGami Robot  
}

\author{Archie Webster, Lee Skull and Seyed Amir Tafrishi*
\thanks{Archie Webster, Lee Skull and Seyed Amir Tafrishi are with Geometric Mechanics and Mechatronics (gm$^2$R) Lab, the School of Engineering, Cardiff University, Cardiff, CF24 3AA, United Kingdom ( e-mail:
        {\tt\small \{WebsterA2, SkullL, Tafrishisa\}@cardiff.ac.uk})}
\thanks{*Seyed Amir Tafrishi is the corresponding author of this study, e-mail: {\tt\small Tafrishisa@cardiff.ac.uk}).}
\thanks{*This work was supported by the Royal Society research grant under Grant \text{RGS\textbackslash R2\textbackslash 242234}.}}
\renewcommand{\baselinestretch}{.94}
\begin{document}

\maketitle
\thispagestyle{empty}
\pagestyle{empty}


\begin{abstract}
This paper presents the design of a new soft-rigid robotic platform, "GeoGami". We leverage origami surface capabilities to achieve shape contraction and to support locomotion with underactuated forms. A key challenge is that origami surfaces have high degrees of freedom and typically require many actuators; we address repeatability by integrating surface compliance. We propose a mono-actuated GeoGami mobile platform that combines origami surface compliance with a geometric compliant skeleton, enabling the robot to transform and locomote using a single actuator. We demonstrate the robot, develop a stiffness model, and describe the central gearbox mechanism. We also analyze alternative cable-driven actuation methods for the skeleton to enable surface transformation. Finally, we evaluate the GeoGami platform for capabilities, including shape transformation and rolling. This platform opens new capabilities for robots that change shape to access different environments and that use shape transformation for locomotion.
\end{abstract}

\section{Introduction}
Compliant origami offers a rich geometric design space where complex 3D behaviors emerge from simple, repeatable folds of a planar sheet \cite{misseroni2024origami,yan2023origami}. Translating these capabilities into robots that achieve shape change or locomotion with minimal actuation is challenging: underactuated designs must exploit structural physics—distributed stiffness, mass and inertia, and geometry-induced constraints—to route energy and motion effectively. This motivates co-design of crease patterns and compliance so the desired transformations can be achieved and controlled with as little actuation as possible.
 
Compliance and softness are crucial for safer, more adaptable mechanisms \cite{yan2023origami} and robotic systems \cite{rus2015design}, with clear advantages in constrained or hazardous settings. In power-plant inspection and maintenance, where intrusive methods are costly and awkward \cite{wang2021robots,10313930}, a robot that can morph to the environment like a soft system yet endure harsh conditions like a rigid platform enables in-situ traversal and stable, on-site operations \cite{liu2021review}. Recent reviews emphasize that compliance must be co-designed with geometry, materials, and control to achieve predictable interaction—spanning modeling frameworks for soft robots \cite{armanini2023softmodeling}, compliant mechanisms for contact-rich tasks \cite{samadikhoshkho2024review}, and system-level overviews of soft robotics \cite{yasa2023overview}. In parallel, origami-based robots illustrate how compliant folding structures integrate sensing/decision/actuation for robust interaction \cite{yan2023origami,seo2019modular}, and surveys document diverse approaches for crease-pattern design and controllable deployment \cite{meloni2021engineeringorigami} or even using geometric one-print compliant models \cite{krysov2024development}. In practice, patterns such as Miura-ori and waterbomb combined with compliant hinges can deliver large, coordinated shape change with minimal actuation, while tendon-, pneumatic-, or SMA-based actuation governs deployment and force transmission. Nevertheless, striking the right balance between rigidity (load-bearing, durability), softness (safe contact, adaptability), and the repeatable “returnability” of origami folds remains non-trivial in practice.

Underactuated robotics, systems with fewer independent actuators than degrees of freedom, offers advantages in energy efficiency, material use, and compact form factors \cite{he2019underactuated,liu2020survey,spong1998underactuated}. When paired with careful morphology and control, such designs can be exceptionally effective in tasks like locomotion \cite{xu2024tracked} and shape transformation in constrained environments \cite{firouzeh2015under}. Because fewer actuators reduce mass, volume, and power draw, underactuated platforms are attractive for field deployment \cite{he2019underactuated}. This mirrors biological strategies, where passive dynamics and compliant couplings distribute motion and forces efficiently. The design goal, therefore, is to use the minimum number of actuators while exploiting mechanical couplings, tendon routing, and structural compliance to modulate stiffness and geometry for navigation and manipulation—an idea exemplified by adaptive-synergy hands coordinating many joints with a single input \cite{catalano2014softhand}. Nevertheless, integrating geometric compliance (e.g., origami-inspired structures) with underactuation to achieve reversible shape change and size modulation using very few actuators remains an open challenge.
\begin{figure}[t!]
      \centering
      \includegraphics[width = 3.4 in]{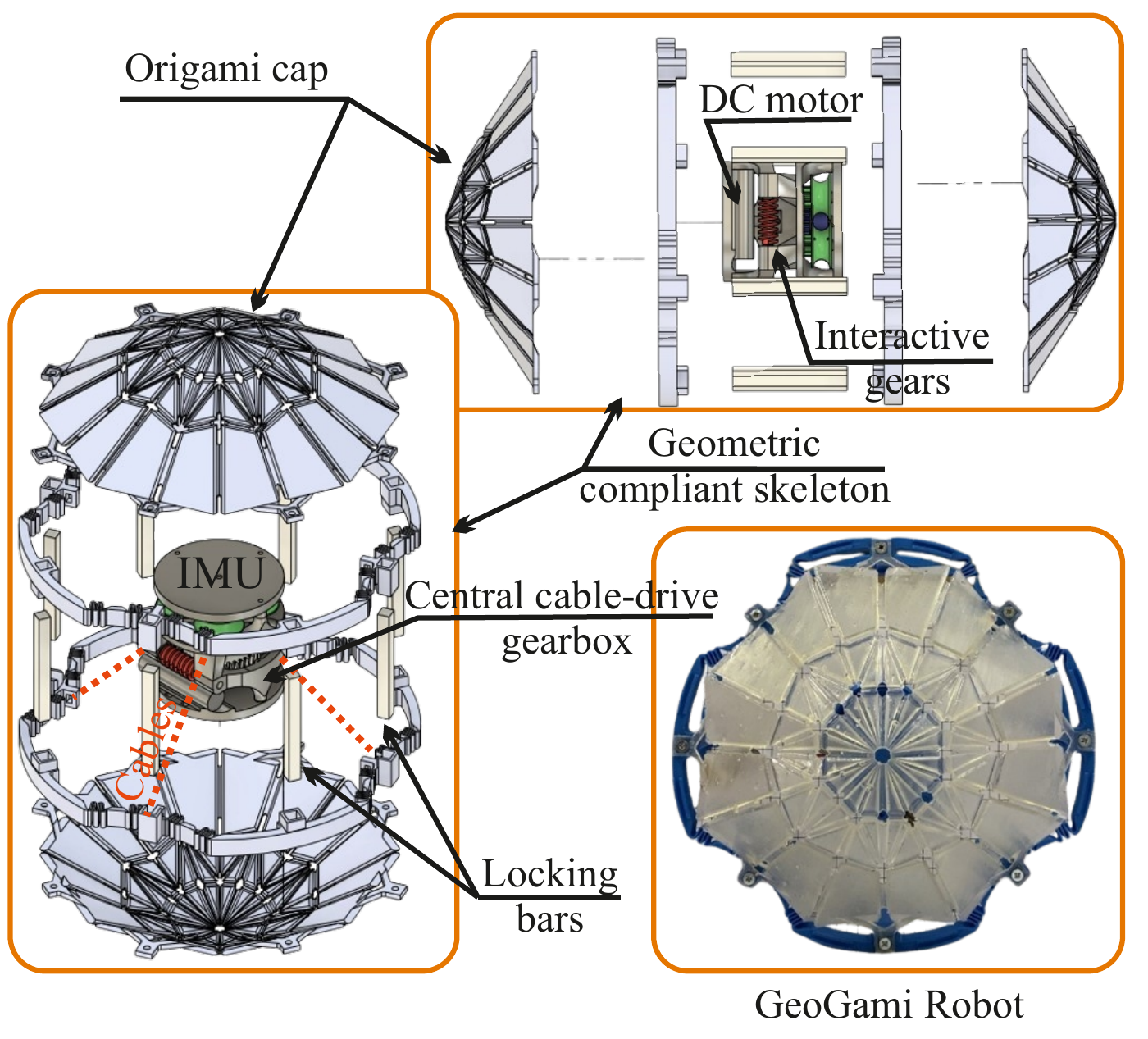}
       \caption{GeoGami robot design and components.}
      \label{fig:GeoGami_Robot}
\end{figure}

We address the limited study of actuation strategies for compliant origami surfaces, particularly when the platform must realize multiple shape transformations in addition to rolling. We introduce novel \emph{GeoGami} as shown in Fig. \ref{fig:GeoGami_Robot}, a mono-actuated robotic platform that integrates a modular geometric compliant skeleton with an origami surface; both structures provide programmable compliance and are manipulated by a centralized cable-driven gearbox. The paper develops and validates a stiffness model for the origami patches and the geometric skeleton, quantifies returnability, and presents an analytical center-of-mass formulation that maps cable retraction through the transmission to mass imbalance and locomotion. We further characterize the transmission and cable cyclic routing for single-actuator operation, and evaluate the resulting motion capabilities of the platform, including size compaction, controllable contact, and rolling.
 
The remainder of the paper is organized as follows. Section~\ref{sec:design} introduces the GeoGami hardware, including the compliant skeleton, origami surface, joint measurement protocol, and the mono-actuated cyclic gearbox. Section~\ref{sec:model} develops the planar kinematics, the stiffness aggregation from joint to side, and the center of mass model that links cable retraction to motion. Section~\ref{sec:motion} presents experiments and motion behavior analysis; Section~\ref{sec:conclusion} concludes findings and outlines future work.


\section{GeoGami Robot Design}
\label{sec:design}
\subsection{GeoGami Body}
The GeoGami body combines a compliant geometric skeleton with an origami shell as shown in Fig.~\ref{fig:GeoGami_Robot}. We derive an equivalent stiffness model, comprising crease torsion and panel bending, to estimate required actuation forces and to support the subsequent kinematic and contact modeling.

\begin{figure}[t!]%
    \centering
    \includegraphics[width=3.2 in]{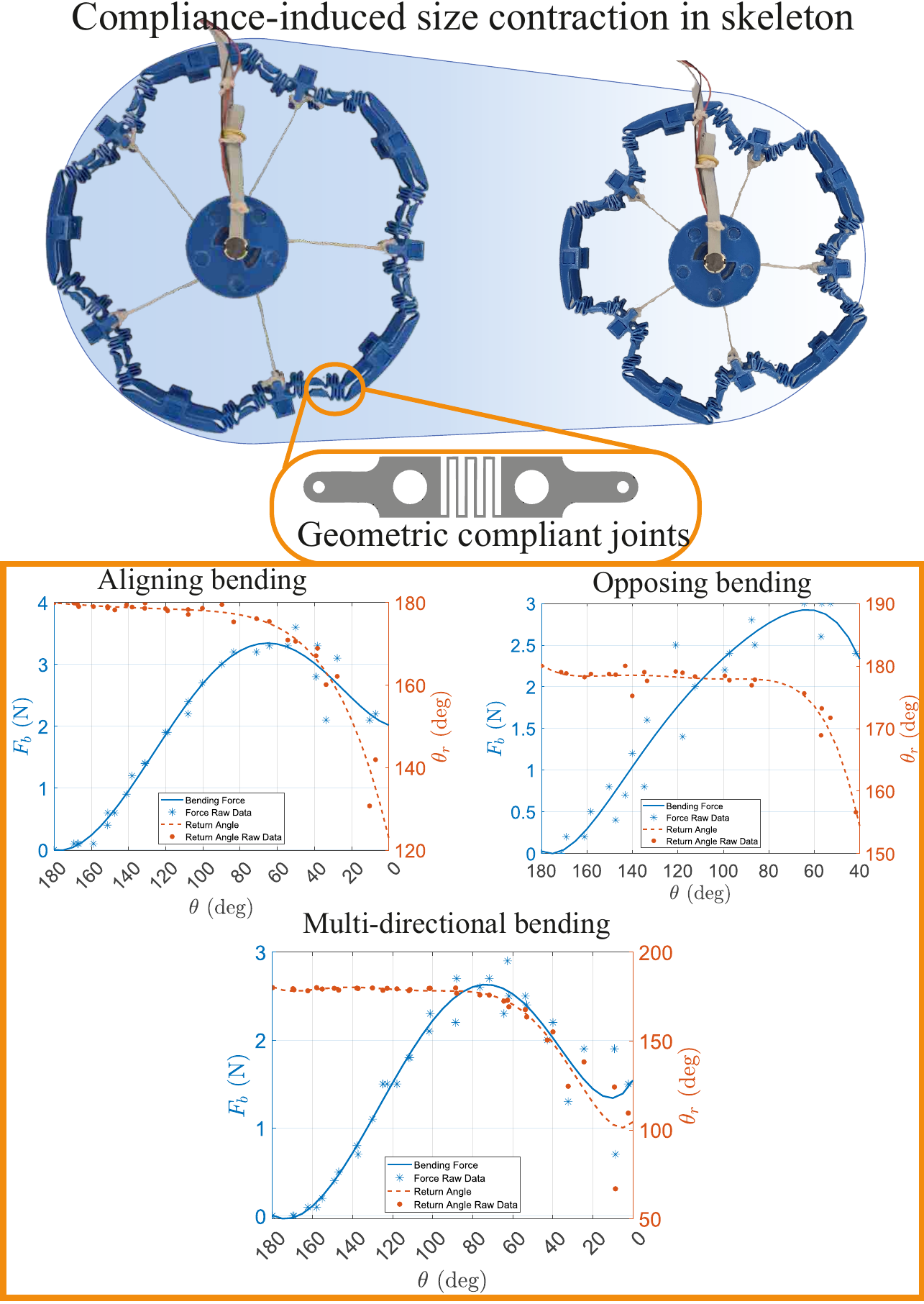} 
    \caption{Compliant geometric skeleton: central cable induces size contraction; bottom plots show \(F_b(\theta)\) and \(\theta_r\) for aligning, opposing, and multi-directional joints.}
    \label{fig:straightbarjoint51}
\end{figure}

The GeoGami robot is designed with two principal parts: (a) a geometric compliant skeleton and (b) a soft compliant origami shell (Fig.~\ref{fig:GeoGami_Robot}). The geometric compliant skeleton is realized as a single 3D-printed PLA lattice that permits controlled deformation under central tendon actuation, as illustrated in Fig.~\ref{fig:straightbarjoint51}. We conducted extensive studies to configure this structure and, in this work, adopt a circular ring with four geometric joints, with two joints positioned on each side of the cable-driven paths that are connected through the central unit. This layout enables shape contraction, as demonstrated in the example shown in Fig.~\ref{fig:straightbarjoint51}.
 
\begin{figure}[t!]%
    \centering
    \includegraphics[width=3.4 in]{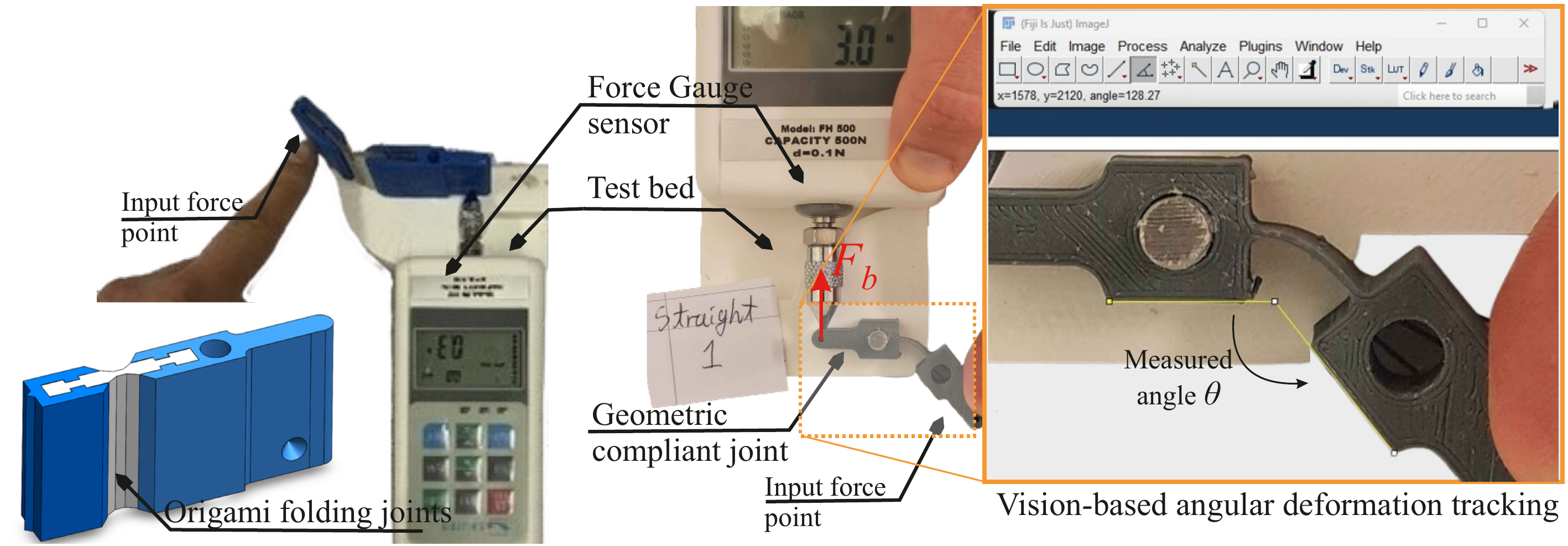} 
    \caption{The measurement of each joint for origami and geometric skeleton for finding stiffness and return angle profiles.}
    \label{measurement_Joint_straight}
\end{figure}


Results as a graph in Fig.~\ref{fig:straightbarjoint51} outline how the compliant geometric skeleton is constructed and characterized for the GeoGami robot. The ring comprises rigid links connected by geometric compliant joints fabricated in PLA; a central cable drive pulls the spokes simultaneously, producing compliance-induced size contraction. Each joint is modeled by its bending force–angle relation \(F_b(\theta)\) measured on a standardized rig, and the angle-dependent bending stiffness is defined as \(k_b(\theta)=\partial F_b/\partial\theta\) which Fig. \ref{measurement_Joint_straight} shows how we do the measurements using Gauss force. For ring-level analysis, joints along a deformation path act approximately in series, giving an effective stiffness \(k_{\!i}(\theta)=\big(\sum_i k_{b,i}^{-1}(\theta)\big)^{-1}\). Recovery is quantified by the elastic return angle \(\theta_r(\theta)\) after unloading. Averaged measurements (markers) with polynomial fits (curves) for three joint families. Aligning joints are the stiffest, yielding large \(F_b\) at moderate bends but a declining \(\theta_r\) as \(\theta\) increases. Opposing joints produce lower peak force while maintaining an almost constant \(\theta_r\) over most of the range. The multi-directional (asymmetric, square-wave–inspired) joint exhibits near-linear \(F_b\)–\(\theta\) behavior up to about \(\theta\!\approx\!70^{\circ}\), supports deep bending with low actuation force, and preserves a high \(\theta_r\); self-contact begins only near \(\theta\!\approx\!150^{\circ}\) and does not impede recovery, and at \(\theta\!\approx\!40^{\circ}\) it retains a larger \(\theta_r\) than symmetric designs. These trends allow tuning of stiffness and returnability to achieve repeatable size contraction with minimal actuation.

\begin{figure}[t!]%
    \centering
    \includegraphics[width=3.2 in]{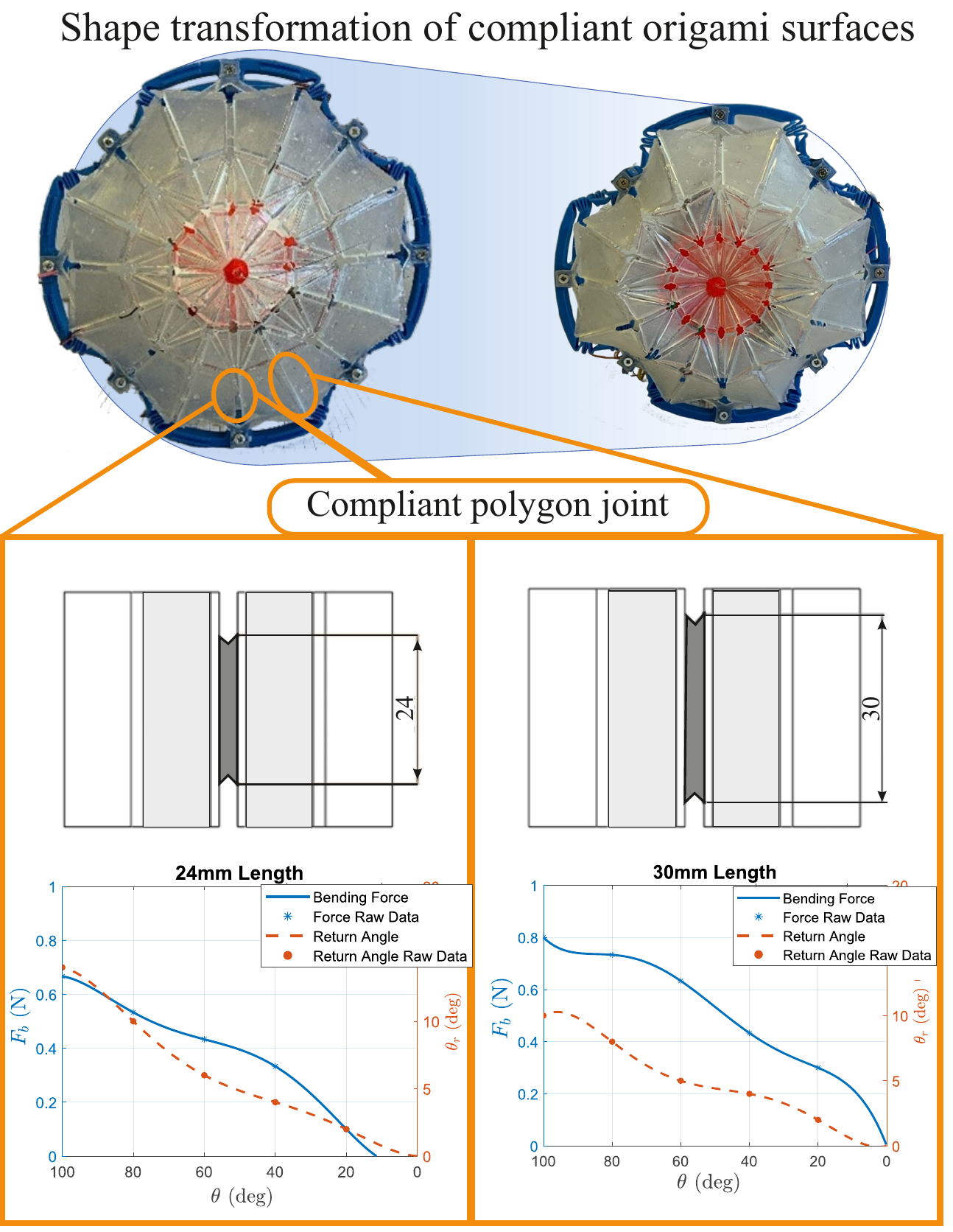} 
    \caption{Compliant origami surface: central cable induces size contraction; bottom plots show \(F_b(\theta)\) and \(\theta_r\) for various folding joints on polygons}
\label{fig:origamisurfacediesgn11}
\end{figure}
\begin{figure*}[t!]
      \centering
      \includegraphics[width = 6.2 in]{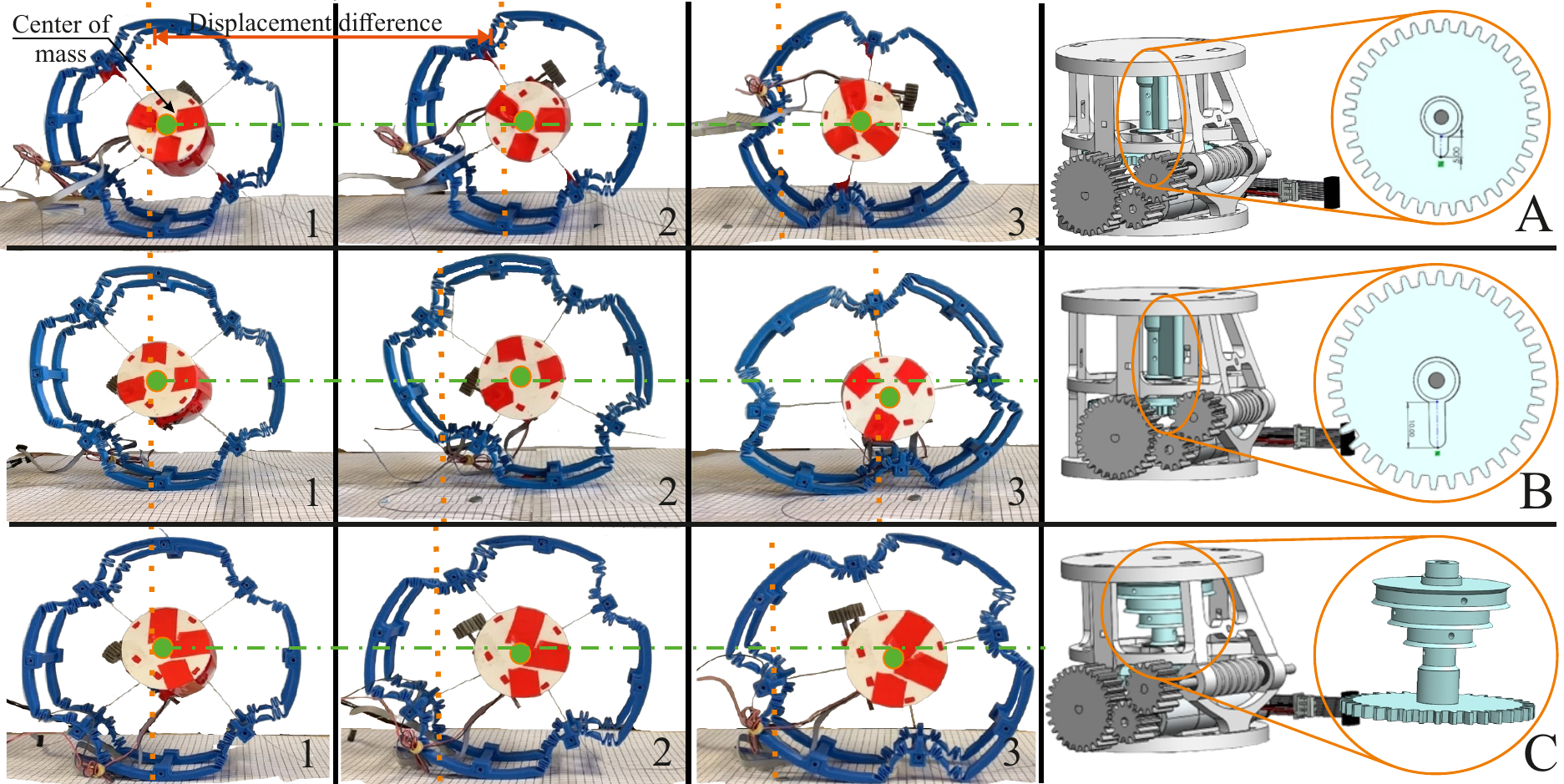}
      \caption{Cable-driven spindle concepts used to deform the geometric skeleton: (A) 5\,mm extruded spindle, (B) 10\,mm extruded spindle, and (C) pyramid spindle.}
      \label{fig:spindle_coupling_motion}
\end{figure*}
The origami surface in Fig.~\ref{fig:origamisurfacediesgn11} is fabricated from a flexible TPU-like photopolymer using SLA, providing reusable folds with moderate compliance. Each crease is implemented as a compliant polygon joint consisting of a thin ligament and relief (air) gaps that prevent local buckling and allow large rotations under cable actuation. We characterized two joint lengths (24\,mm and 30\,mm) and thickness of 2.4 mm width using the same bending protocol as for the skeleton: the measured force–angle curves \(F_b(\theta)\) decrease approximately monotonically as \(\theta\) closes, with \(F_b\!\approx\!0.7\!-\!0.9\) N near \(\theta\!\approx\!100^\circ\) and tending toward zero as \(\theta\!\to\!0^\circ\); the elastic return angle \(\theta_r\) drops from about \(15\!-\!18^\circ\) at large bends to near \(0^\circ\) at small \(\theta\). The longer 30\,mm joint exhibits a lower stiffness profile than the 24\,mm joint, consistent with increased compliant length. Since these measurements are for a single crease, the effective force at a cable anchor on the full cap scales with the number of engaged polygon joints (five per sector in our design), which combine approximately in parallel. This joint geometry provides sufficient flexibility for repeated shape change while preserving predictable recovery of the origami surface.

\subsection{Mono-Actuated Gearbox}
\begin{figure}[t!]
    \centering
    \includegraphics[width=\linewidth]{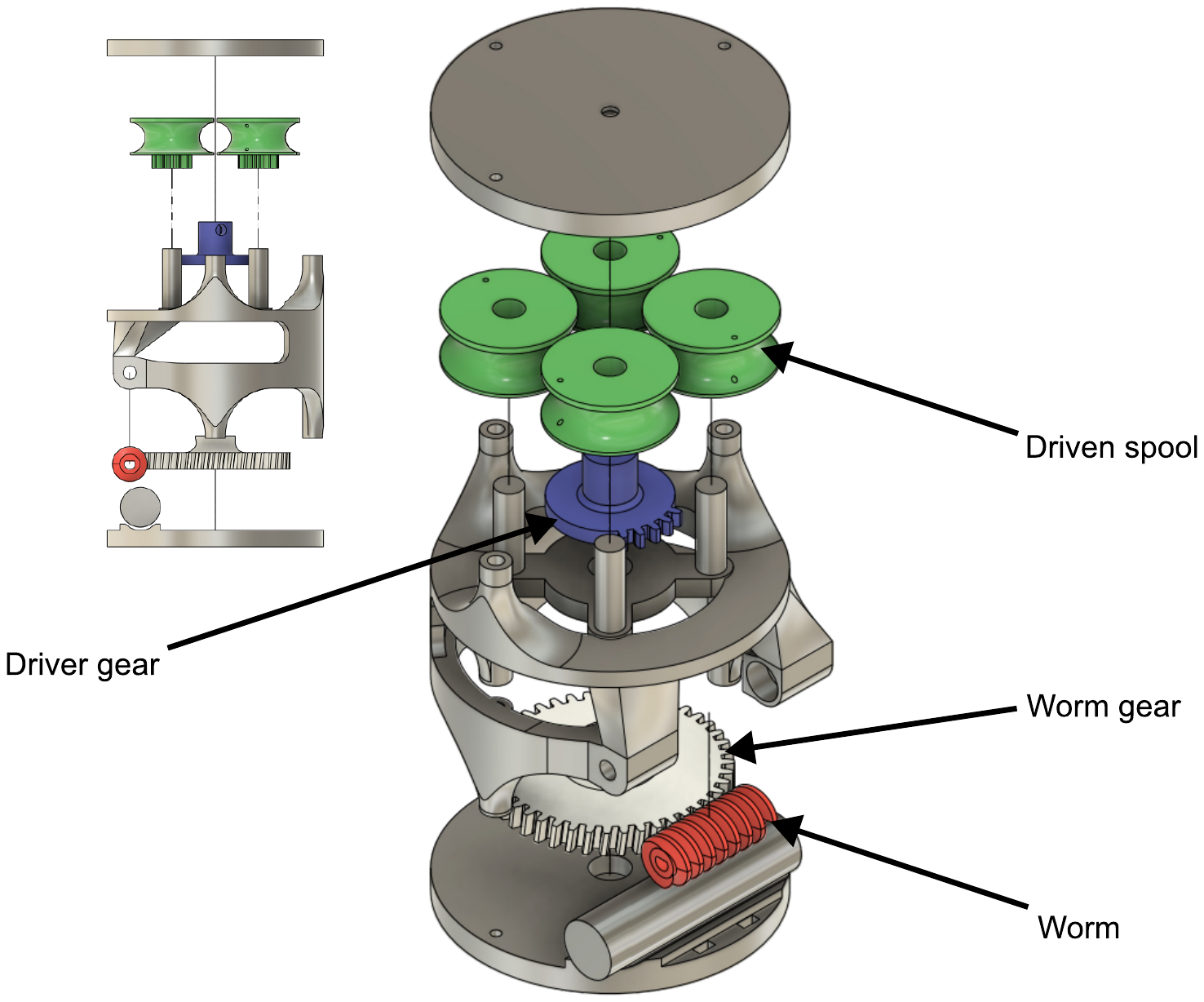}
    \caption{Cyclic cable-drive gearbox that time-multiplexes pulls to the four corners.}
    \label{fig:annotatedExplodedcyclic}
\end{figure}

The target design is a centralized cable-driven gearbox that allows actuation of compliant skeletons of different sizes, together with origami caps, to enable both shape change and locomotion. We begin by evaluating three cable-driven spindle designs (Fig.~\ref{fig:spindle_coupling_motion}): a pyramid spindle, a 5\,mm extruded spindle, and a 10\,mm extruded spindle. All induce differential cable take-up that deforms the geometric skeleton asymmetrically and displaces the center of mass, producing repeatable tipping and approximately \(45^\circ\) rolling. The pyramid spindle contracts sectors at different rates as wire engages larger diameters; it provides reliable rocking with full return when polarity is reversed, but cannot sustain continuous rotation because contraction saturates and the support polygon collapses to a tripod that arrests motion. The 5\,mm spindle introduces a mild azimuthal phase in contraction; it also attains \(45^\circ\) and reconfigures, yet remains slow due to the small diameter and still locks into a tripod contact that prevents full rotation. The 10\,mm spindle increases local cable retraction per revolution, generating a faster center-of-mass shift and earlier toppling; it reaches \(45^\circ\) quickly but cannot unwind to the initial shape, leaving the robot immobile at maximum contraction. In summary, fixed spindle geometries create large but localized asymmetry without a traveling phase around the ring; once maximum contraction is reached, the center of mass cannot advance, so continuous locomotion is inefficient or impossible. We therefore adopt a cycle-based actuation strategy that time multiplexes cable pulling around the perimeter to generate a traveling centre-of-mass displacement and sustained rolling.

The proposed another gearbox (Fig.~\ref{fig:annotatedExplodedcyclic}) uses a single–start worm on the motor shaft that drives a worm gear rigidly coupled to a vertical driver shaft. A sector (mutilated) gear mounted on this shaft intermittently engages each corner’s driven gear–spool set, so one motor sequentially retracts the four cables. Each spool is attached to a corner of the skeleton with a cable. When the spool is driven, it rotates, winding in the cable and contracting one corner of the skeleton. When the driver gear's teeth have passed the spool, it disengages and the cable is unwound by the skeleton returning back to the initial position (full radius length of cable). The worm gear allows the skeleton to hold its retracted shape if required, as it cannot be back-driven. This allows the mechanism to be programmable and controllable to activate neighboring cables depending on the position of the COM in the rolling GeoGami robot or contact-based shape change.

To compute amount of displacement happen in cables, at first we assume angles are in radians and taken positive when winding. We assume a pretensioned, single-layer wrap (no slip). Let \(\theta_m\) be the motor angle, a single–start worm drives a worm gear
with \(T_w\) teeth; hence the driver–shaft angle is
\begin{equation}
\theta_d \;=\; \frac{\theta_m}{T_w}.
\label{eq:worm_gbx}
\end{equation}
When the sector (mutilated) gear on the driver shaft is in mesh with
corner \(i\) (engagement indicator \(\chi_i=1\); otherwise \(\chi_i=0\)),
the driver gear with \(T_{dr}\) teeth drives the corner’s spur gear with
\(T_{dv}\) teeth, and the coaxial spool rotates by
\begin{equation}
\theta_{s,i} \;=\; \chi_i\,\theta_d\frac{T_{dr}}{T_{dv}}
               \;=\; \chi_i\,\theta_m\frac{T_{dr}}{T_{dv}\,T_w}.
\label{eq:sector_gbx}
\end{equation}
For a spool of radius \(r_s\), the cable length change at corner \(i\) is
\begin{equation}
L_i \;=\; r_s\,\theta_{s,i}.
\label{eq:arcLen_gbx}
\end{equation}
Finally, we can find the motor rotation needed to retract a prescribed length
\(L_i\) during engagement is
\begin{equation}
\theta_m \;=\; \frac{L_i\,T_{dv}\,T_w}{\chi_i\,r_s\,T_{dr}}.
\label{eq:motor_from_L_gbx}
\end{equation}

The sector’s toothed arc subtends \(\alpha\) radians, so each corner is
engaged for a duty factor \(D=\alpha/(2\pi)\). With constant motor speed
\(\dot\theta_m\), the engagement advances around the ring with phase
velocity
\begin{equation}
\omega_{\text{phase}} \;=\; D\,\frac{\dot\theta_m}{T_w}.
\label{eq:phase_velocity}
\end{equation}

Using the kinematic relations in \eqref{eq:worm_gbx}–\eqref{eq:arcLen_gbx}, the torque transmitted to the spool at corner \(i\) is amplified by the worm and spur stages. Let \(\eta_w\) and \(\eta_g\) denote the efficiencies of the worm stage and the driver–driven spur pair, with total efficiency \(\eta=\eta_w\eta_g\in(0,1)\), and \(\tau_m\) the motor torque. Neglecting inertia (quasi static), the spool torque is
\begin{equation}
\tau_{s,i} \;\triangleq\; \chi_i\, \eta \, T_w \,\frac{T_{dv}}{T_{dr}}\, \tau_m .
\label{eq:spool_torque}
\end{equation}
With a single–layer wrap on a spool of radius \(r_s\), the cable tensile force is determined
\begin{equation}
F_i \;= \; \frac{\tau_{s,i}}{r_s}
      \;=\; \chi_i\, \frac{\eta\, T_w\, T_{dv}}{T_{dr}\, r_s}\,\tau_m .
\label{eq:cable_force}
\end{equation}


\section{Model of GeoGami} 
\label{sec:model}
\begin{figure}[t!]
      \centering
      \includegraphics[width = 0.45\textwidth]{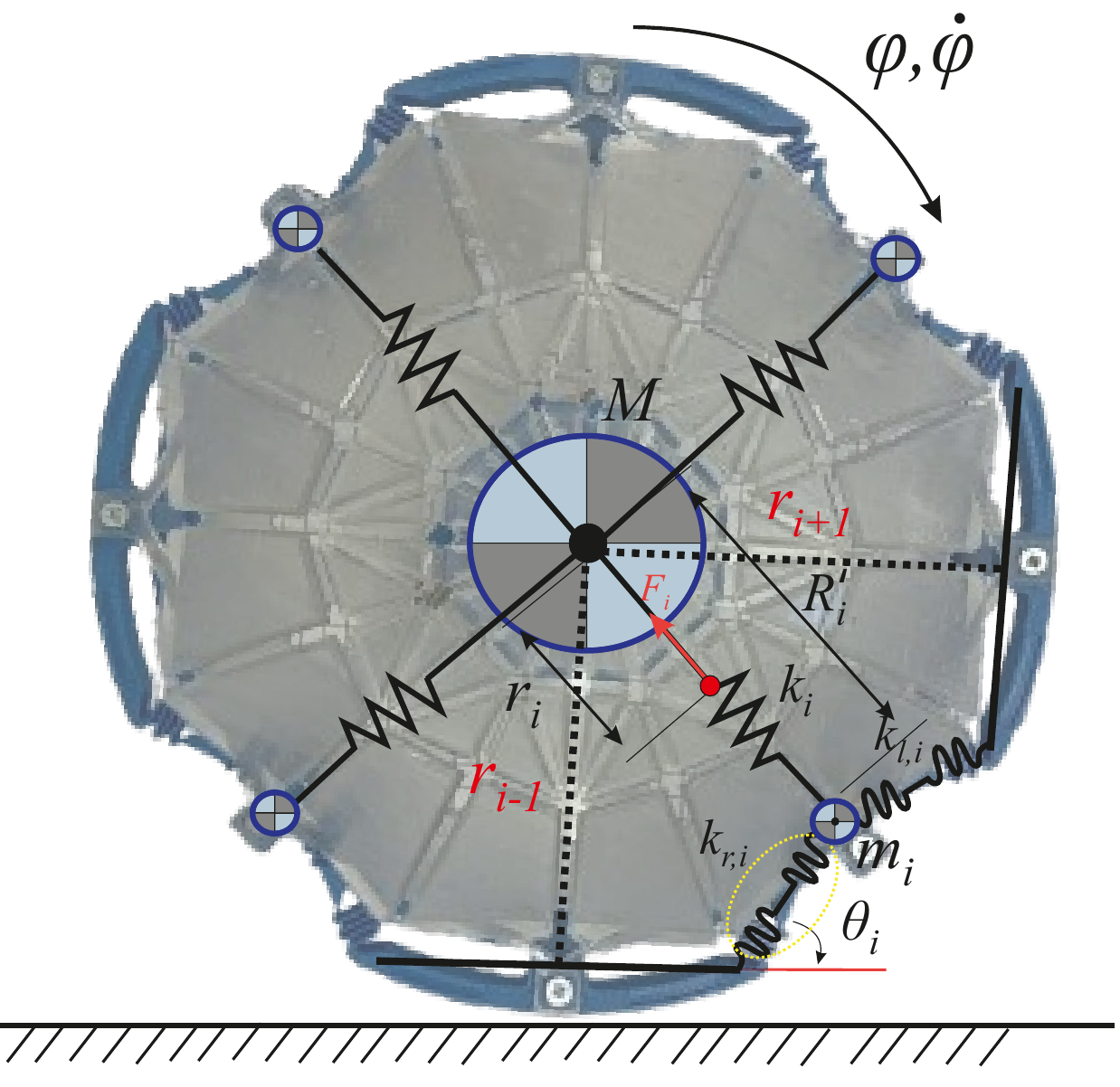}
      \caption{Schematic of spring model of origami with mass distribution.}
      \label{fig:LEE_Geometric}
\end{figure}

This section formalizes the planar kinematics of the rolling GeoGami platform, stiffness model and the computation of the center of mass (COM) that links cable–drive actuation to rolling through mass–imbalance based on Fig. \ref{fig:LEE_Geometric}. The origami cap is approximated by four radial spring–linkages with mass-point attached as $m_i$ and central gearbox mass as $M$. On side \(i\), the net radial stiffness is denoted \(k_i\) and is realized by two compliant segments with left and right stiffnesses \(k_{l,i}\) and \(k_{r,i}\). The instantaneous radial distances from the geometric center to the four link endpoints are \(r_i(t)\), which are functions of the cable retractions routed to each side (see Fig.~\ref{fig:LEE_Geometric}). We adopt the rolling–without–slip coordinate \(\varphi(t)\) and the support radius \(R(t)\). The translation of the GeoGami's body center $\uvec{r}_s$ and the offset point used to describe the mass–imbalance $\uvec{r}_o$ are
\begin{align}
\uvec{r}_{s} =  \begin{bmatrix}  R(t) \varphi(t) \\
0\end{bmatrix}  ,
\uvec{r}_{o} =     \begin{bmatrix}
        R(t) \varphi+  (r_2(t)-r_4(t)) \cos{\varphi} \\
        - (r_1(t)-r_3(t)) \sin{\varphi}
    \end{bmatrix}  
    \label{EQ:positionvector}
\end{align}
where the horizontal pair \((r_2,r_4)\) contributes to the \(x\)–offset through \(\cos\varphi\) and the vertical pair \((r_1,r_3)\) contributes to the \(y\)–offset through \(\sin\varphi\) with the shown sign convention. Differentiating \eqref{EQ:positionvector} gives the corresponding velocities,
\begin{align*}
&\dot{\uvec{r}}_{s} =  \begin{bmatrix}  \dot{R}\varphi+  R \dot{\varphi}\\
0\end{bmatrix}, \\
&\dot{\uvec{r}}_{o} =     \begin{bmatrix}
       \dot{R}\varphi+  R \dot{\varphi}+  (\dot{r}_2-\dot{r}_4) \cos{\varphi}-\dot{\varphi}(r_2-r_4) \sin{\varphi}\nonumber\\
        - (\dot{r}_1-\dot{r}_3)\sin\varphi+\dot{\varphi} (r_1-r_3) \cos{\varphi}
    \end{bmatrix} .
    \label{EQ:velvector}
\end{align*}

We model a centered gearbox mass \(M\) at the geometric origin and four point masses \(m_i\) collocated as explained with the spring–link endpoints. Each endpoint \(i\in\{1,2,3,4\}\) lies on a body–fixed ray of nominal radius \(R_i\) and angle \(\alpha_i\). Compliance produces a radial shortening \(u_i(t)\), so the instantaneous radius is
\begin{equation}
\label{eq:ri_def}
r_i(t) = R_i - u_i(t).
\end{equation}
To identify the side–wise stiffness, we model the origami cap along each
radial line as a series chain of compliant folds that is lumped into a
single equivalent spring \(k_i(t)\) (i.e., if the chain contains elements
\(k_{i,j}\), then \(k_i^{-1}=\sum_j k_{i,j}^{-1}\)). The geometric compliant
skeleton on the same side contributes two parallel segments with stiffnesses
\(k_{l,i}(t)\) and \(k_{r,i}(t)\). Since all rest elements lie in series
along the pulling direction, the effective \emph{radial} stiffness used for
the contraction \(u_i\) is
\begin{equation}
\label{eq:keff_i}
\kappa_i^{-1} \;=\; k_i^{-1}\;+\left(\;k_{l,i}\;+\;k_{r,i}\right)^{-1},
\end{equation}
where \(\kappa_i\) represents the side-\(i\) equivalent of the origami
patch–skeleton assembly. Under quasi-static loading, the local radial force
satisfies \(F_i(t) \approx \kappa_i(t)\,u_i(t)\).

\begin{figure*}[t!]
  \centering
  \begin{minipage}[b]{0.44\textwidth}
    \centering
    \includegraphics[width=\linewidth]{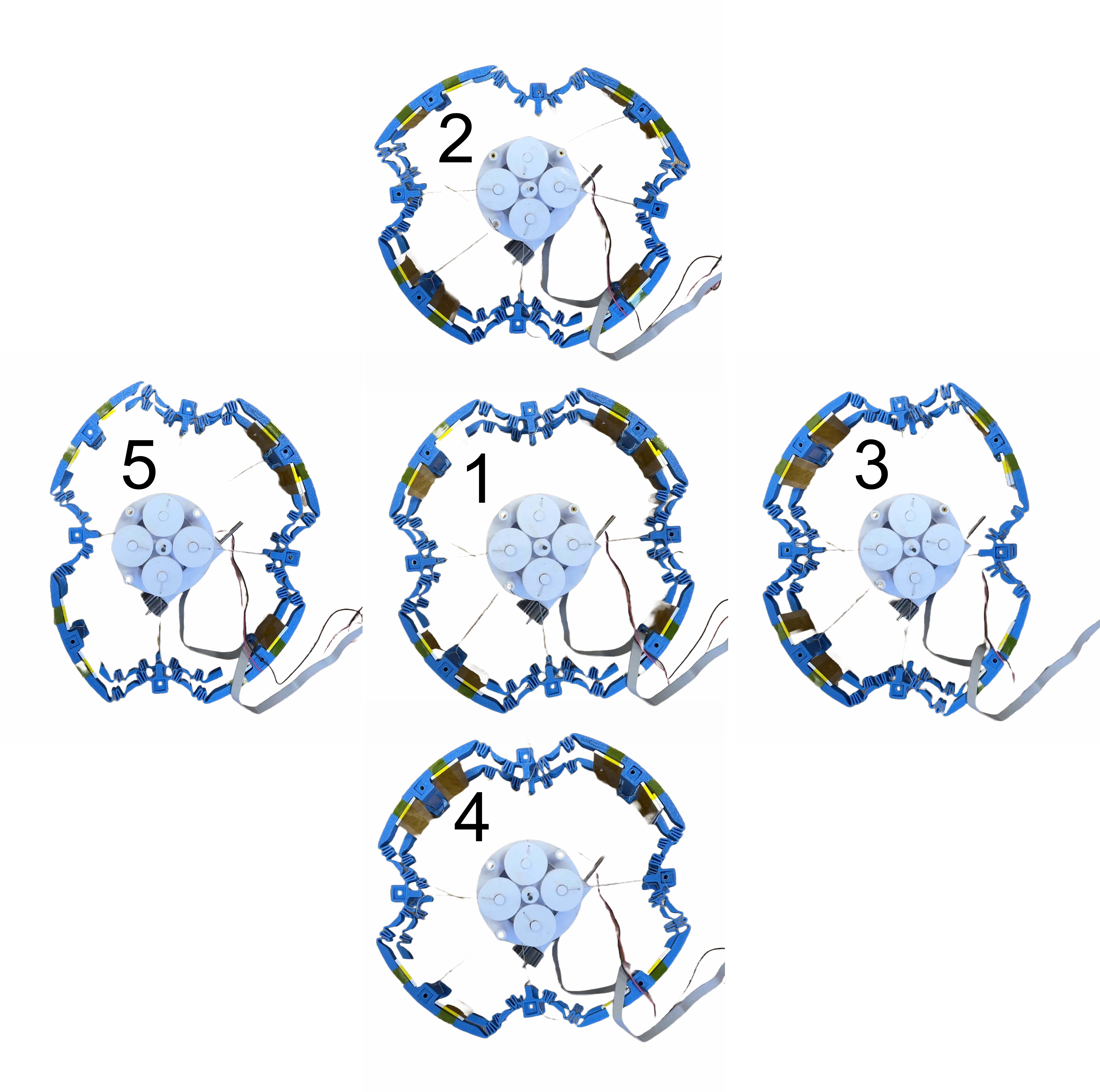}
    \vspace{4pt}
    {\footnotesize \textbf{(a) Without origami cap.} }
  \end{minipage}\hfill
  \begin{minipage}[b]{0.44\textwidth}
    \centering
    \includegraphics[width=\linewidth]{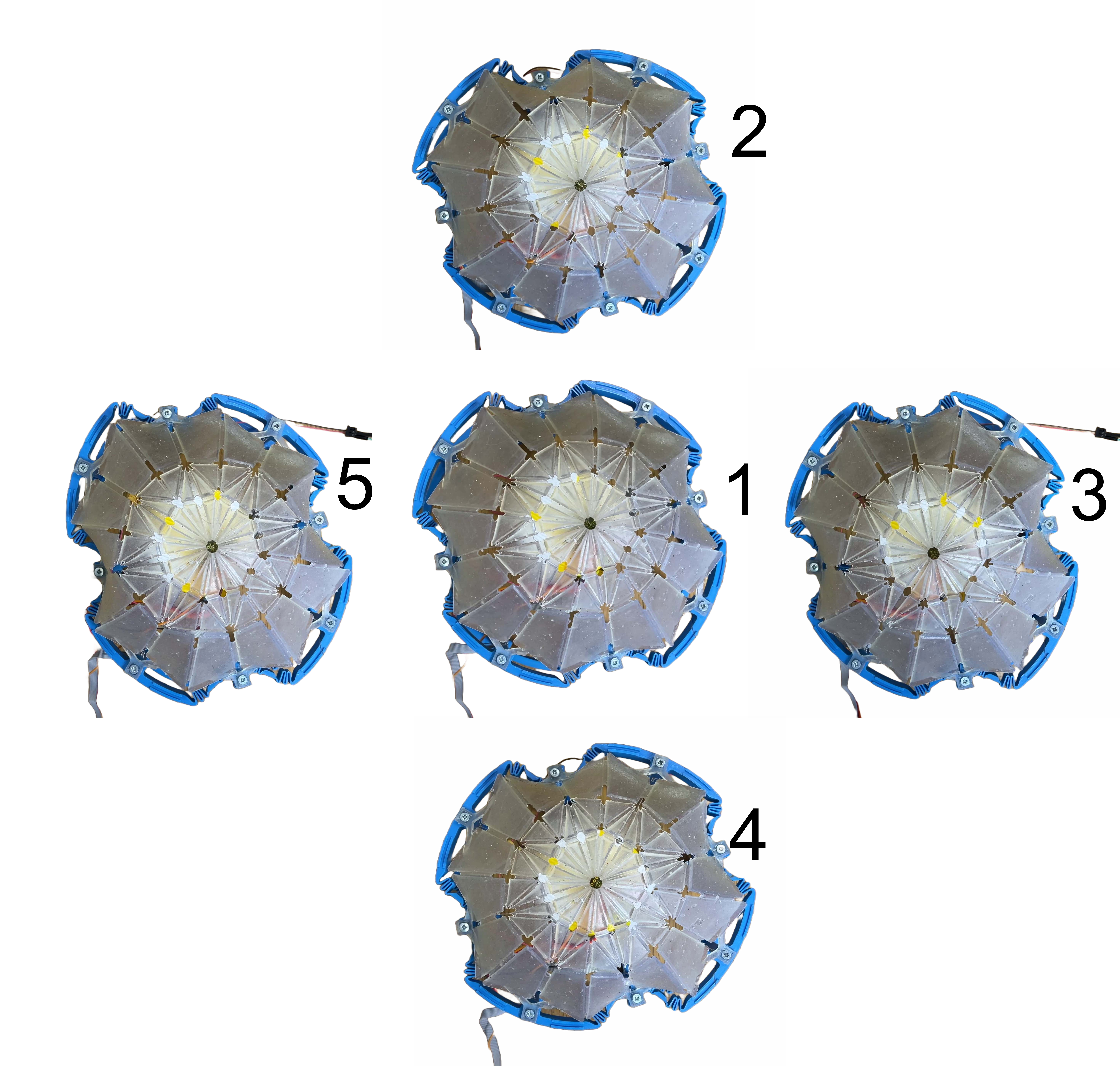}
    \vspace{4pt}
    {\footnotesize \textbf{(b) With origami cap.} }
  \end{minipage}
  \caption{Overhead actuation sequences with and without the origami cap (1: unactuated; 2–5: actuation). The origami increases radial compliance and distributes deformation while preserving the cyclic gearbox timing.}
    \label{fig:noOrigamiOverhead}
\label{fig:WithOrigamiOverhead}
\end{figure*}

Each side is driven by a cable. If the axial cable stiffness on side \(i\)
is \(k_{c,i}\) and the routing converts radial contraction into cable
retraction through a known geometric gain \(g_i\) (so that
\(\Delta\ell_i = g_i\,u_i\), with \(g_i\!\approx\!1\) for a radial path),
the stiffness seen in the cable is the series combination
\begin{equation}
\label{eq:Ki_series}
K_i^{-1}(t) \;=\; \kappa_i^{-1}(t) \;+\; k_{c,i}^{-1},
\end{equation}
and the cable tension is
\begin{equation}
\label{eq:T_i}
T_i(t) \;=\; K_i(t)\,u_i(t) \;=\; \frac{K_i(t)}{g_i}\,\Delta\ell_i(t).
\end{equation}
where the tension \(T_i\) maps to actuator torque (\ref{eq:spool_torque}) through the gearbox ratio and
spool radius as described in the transmission section; together with
\eqref{eq:keff_i}–\eqref{eq:T_i} this connects cable retractions to the COM
shift and rolling kinematics defined in \eqref{EQ:positionvector}. The platform rolls without slip and rotates by \(\varphi(t)\), so the translation of the body center is
\begin{equation}
\label{eq:rs}
\mathbf r_{s}(t)=
\begin{bmatrix} R(t)\,\varphi(t)\\[2pt] 0 \end{bmatrix}.
\end{equation}
The planar rotation matrix is
\begin{equation}
\label{eq:R}
\mathbf R(\varphi)=
\begin{bmatrix}\cos\varphi & -\sin\varphi\\[2pt]\sin\varphi & \cos\varphi\end{bmatrix}.
\end{equation}

To find the COM while rotation, we assume in the body frame, the position of mass \(m_i\) is
\begin{equation}
\label{eq:pi}
\mathbf p_i(t)= r_i(t)
\begin{bmatrix}\cos\alpha_i\\[2pt]\sin\alpha_i\end{bmatrix},
\qquad i=1,\ldots,4 .
\end{equation}
Let the total mass be \(M_T = M + \sum_{i=1}^{4} m_i\) and define the mass–weighted offset in the body frame
\begin{equation}
\label{eq:db}
\mathbf d_b(t)=\frac{1}{M_T}\sum_{i=1}^{4} m_i\,\mathbf p_i(t).
\end{equation}
The center of mass (COM) in the world frame is therefore
\begin{equation}
\label{eq:rG}
\mathbf r_G(t)=
\mathbf r_{s}(t)+\mathbf R\!\big(\varphi(t)\big)\,\mathbf d_b(t).
\end{equation}
For the canonical fourfold layout \((\alpha_1,\alpha_2,\alpha_3,\alpha_4)=(\tfrac{\pi}{2},0,\tfrac{3\pi}{2},\pi)\), \eqref{eq:db} gives
\begin{equation}
\label{eq:db_components}
\mathbf d_b(t)=
\frac{1}{M_T}
\begin{bmatrix}
m_2\,r_2(t)-m_4\,r_4(t)\\[4pt]
m_1\,r_1(t)-m_3\,r_3(t)
\end{bmatrix}.
\end{equation}
Substituting \eqref{eq:db_components} into \eqref{eq:rG} and using (\ref{EQ:positionvector}) yields the component form
\begin{align}
\label{eq:rGxy}
x_G &= R\varphi+\cos\varphi\,d_{bx}-\sin\varphi\,d_{by},\\
y_G &= \sin\varphi\,d_{bx}+\cos\varphi\,d_{by},
\end{align}
with \([d_{bx}\ d_{by}]^\top = \mathbf d_b\). If \(m_1=\cdots=m_4\) and \(M\) is constant, \eqref{eq:rG}–\eqref{eq:db_components} recover the offset structure used in \eqref{EQ:positionvector}, making the COM expression fully consistent with the kinematic model.

\section{Motion Analysis}
\label{sec:motion}
We evaluate the compliant mechanism for both size compaction (shape transformation) and rolling locomotion using the mono-actuated cyclic spool gearbox. To isolate the effect of surface compliance, we report sequences without and with the origami cap (Fig.~\ref{fig:GeoGami_Robot}). The gearbox design variables used in these trials are summarized in Table~\ref{tab:variables}; in all experiments each cable undergoes a retraction of 25.1 mm. With the same geometric skeleton, the cyclic actuation produces a traveling deformation that compacts the structure and shifts the center of mass, enabling underactuated shape change and locomotion from a single input. The folding polygon joints exhibit near-constant bending stiffness over \(0\text{--}1.75~\mathrm{rad}\), with \(\bar{k}_b^{(24)} \approx 0.52~\mathrm{N\,rad^{-1}}\) for the 24\,mm joint and \(\bar{k}_b^{(30)} \approx 0.43~\mathrm{N\,rad^{-1}}\) based on measurements as Fig. \ref{fig:origamisurfacediesgn11}, this results in $k_i$ as 0.096 N/rad with five folding joints. Using the series combination in \eqref{eq:keff_i} with Table~\ref{tab:variables} \((k_i=0.096,\;k_{l,i}=k_{r,i}=0.6)\) and treating these as linearized radial stiffnesses over the operating range, the side–equivalent stiffness is \(\kappa_i=\big(k_i^{-1}+k_{l,i}^{-1}+k_{r,i}^{-1}\big)^{-1}\approx 0.073\), so for one engaged corner \((\chi_i=1)\) and \(g_i\!\approx\!1\) the single–wire tension from \eqref{eq:Ki_series}–\eqref{eq:T_i} with \(L_i=25.1~\mathrm{mm}\) is \(T_i\approx \kappa_i L_i \approx 1.8~\mathrm{N}\).
Mapping this tension to motor torque via \eqref{eq:cable_force} with values given in Table \ref{tab:variables} gives \(\tau_m \approx T_i r_s\, T_{dr}/(\eta T_w T_{dv}) \approx 2.4\times10^{-4}\ \mathrm{N\,m}\) per active wire.

\begin{table}[h]
\centering
\caption{Experimental Parameters}
\label{tab:variables}
\begin{tabular}{l c}
\hline
Parameter & Value \\
\hline
Motor voltage & 6.0 V \\
Spool radius $r_s$ & 8 mm \\
Worm gear teeth $T_w$ & 43 \\
Driver gear teeth $T_{dr}$ & 5 \\
Driven gear teeth $T_{dv}$ & 10 \\
Cable retraction length $L$ & 25.1 mm\\
Effective stiffness origami $k_i$ & 0.096 N/rad. \\
Left and right stiffness geometric skeleton $k_{r,i}/k_{l,i}$ & 0.6 N/rad. \\
\hline
\end{tabular}
\end{table}

\begin{figure}[t!]
      \centering
      \includegraphics[width = 3.4 in]{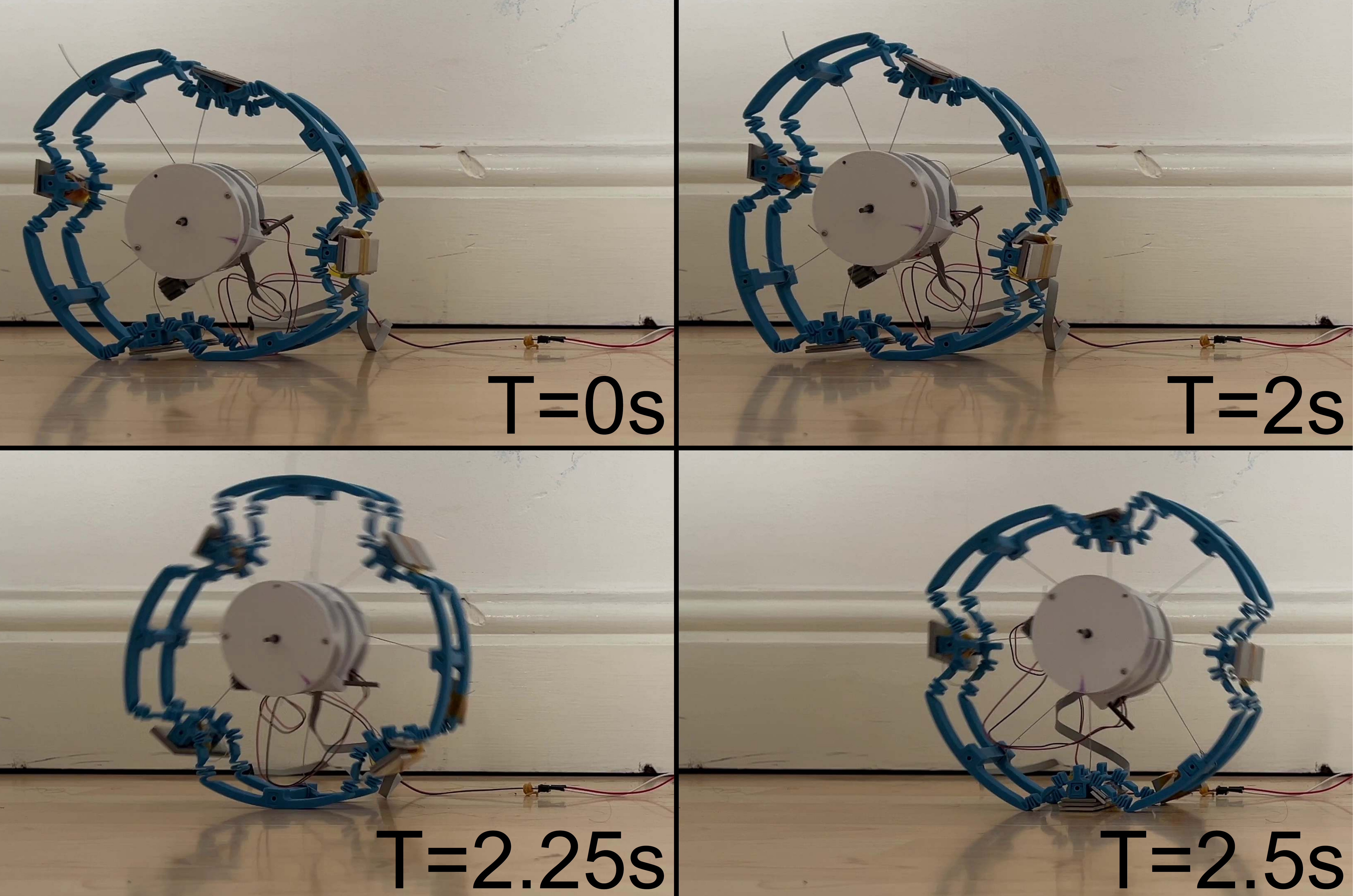}
       \caption{Rolling motion without origami structure (direction of rolling is from left to right).}
      \label{fig:noOrigami4FramesAnnotated}
\end{figure}

\begin{figure}[t!]
      \centering
      \includegraphics[width = 3.4 in]{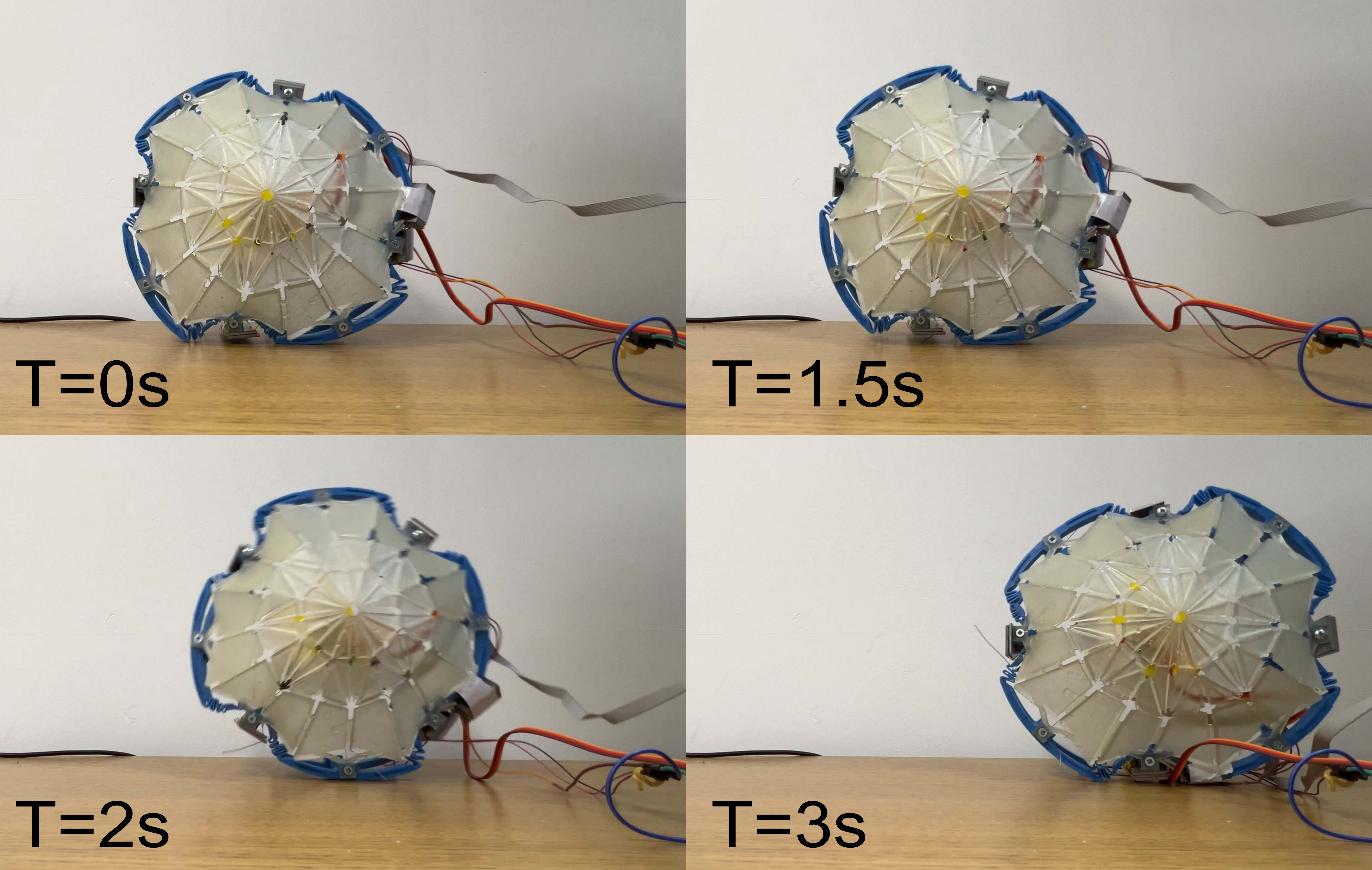}
       \caption{Rolling motion with origami structure.}
      \label{fig:withOrigami4FramesAnnotated}
\end{figure}

\begin{figure}[t!]
    \centering
    \includegraphics[width=0.235\textwidth]{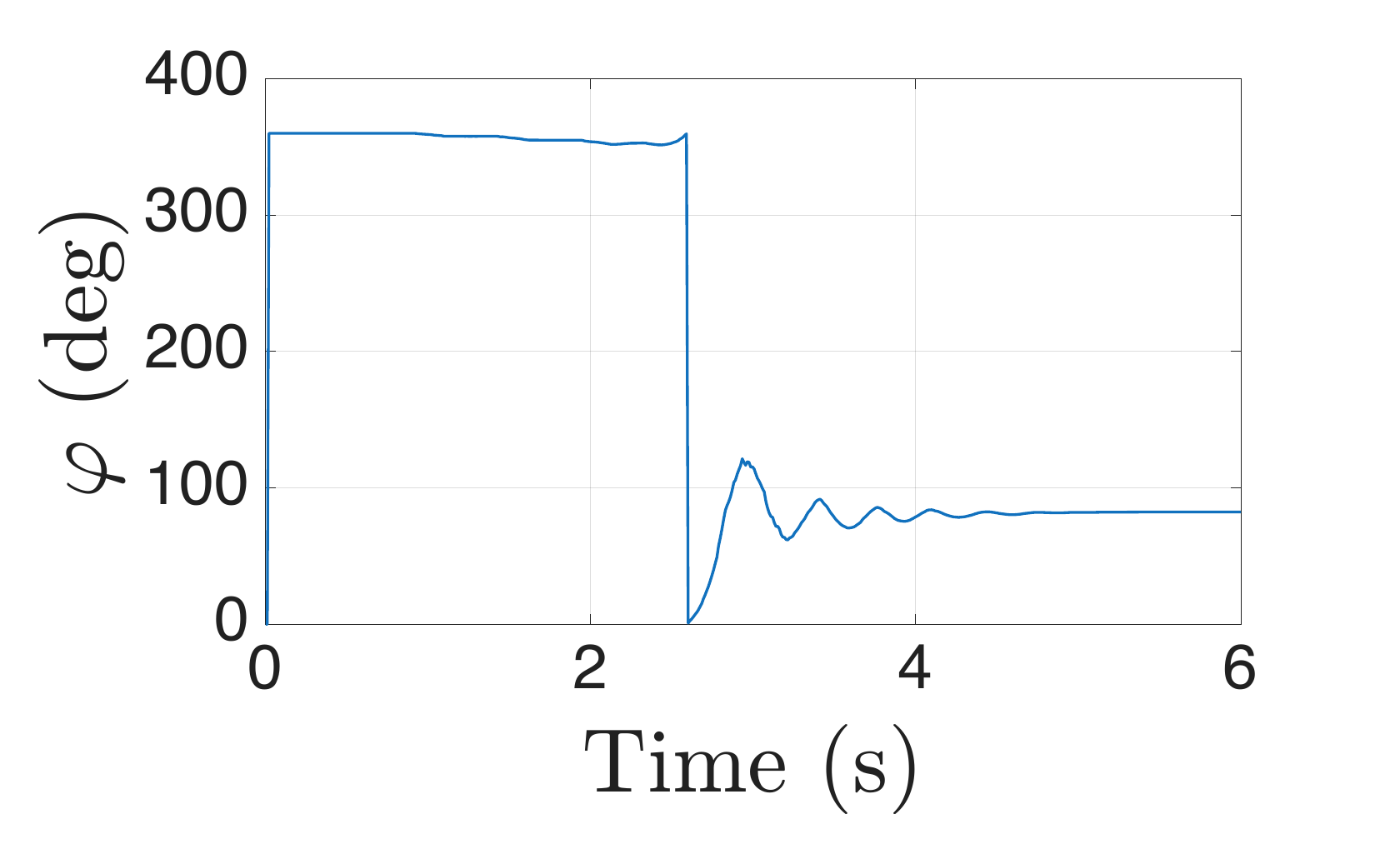}
    \hfill
    \includegraphics[width=0.235\textwidth]{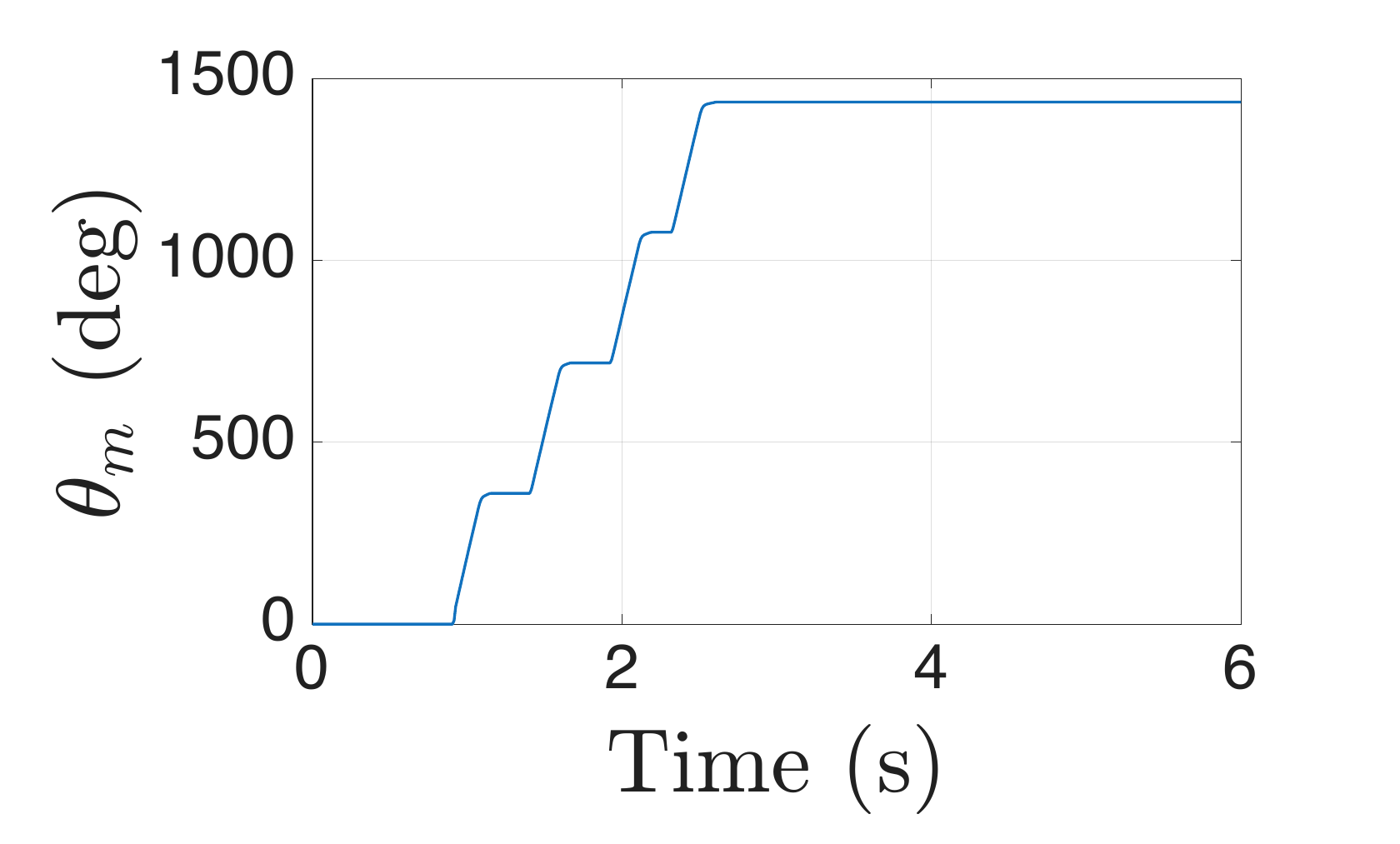}
    \caption{Rotation angle and motor shaft position during rolling movement without origami captured with IMU and motor encoder.}
    \label{fig:IMUdataNoOrigami}
\end{figure}

\begin{figure}[t!]
    \centering
    \includegraphics[width=0.235\textwidth]{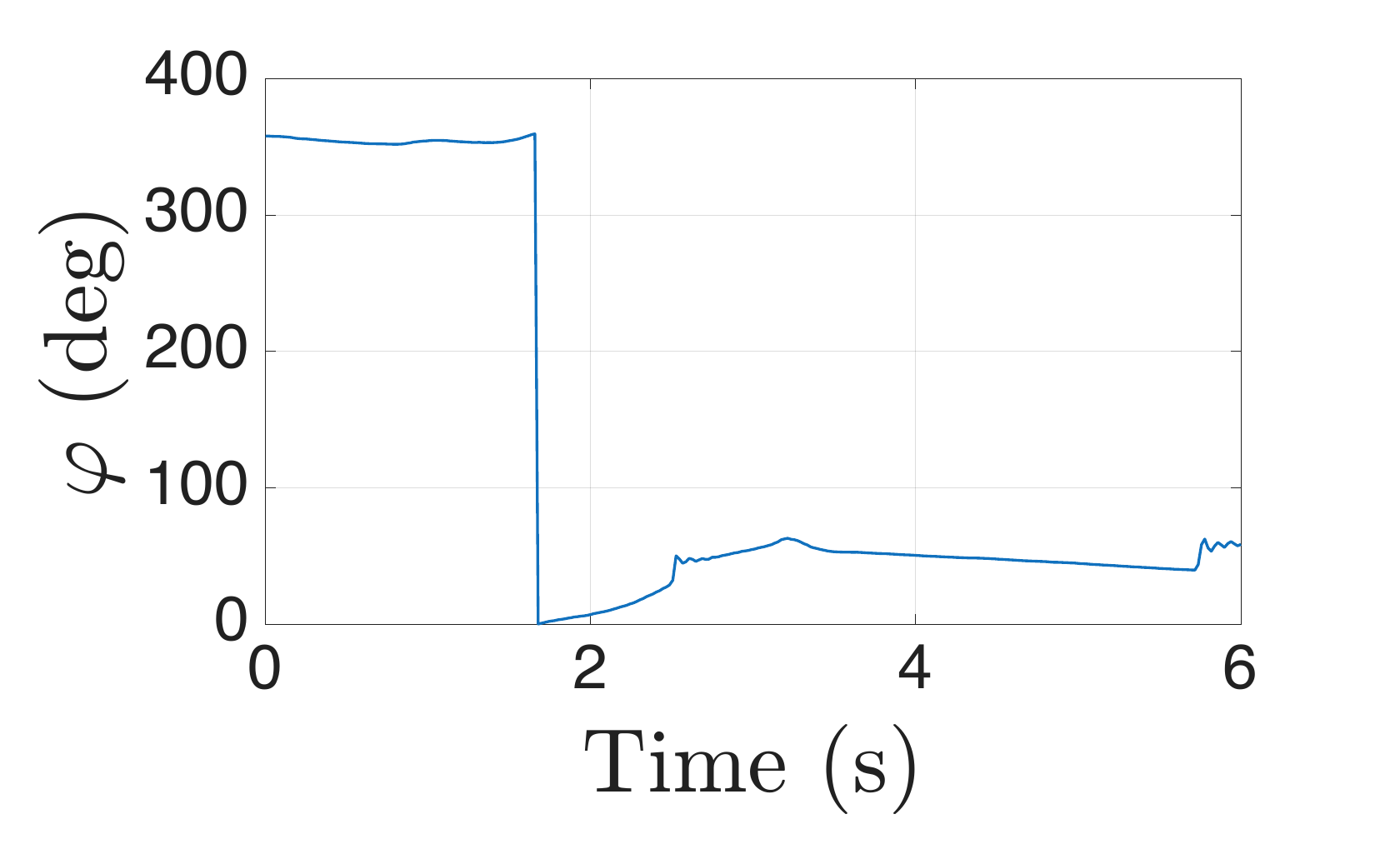}
    \hfill
    \includegraphics[width=0.235\textwidth]{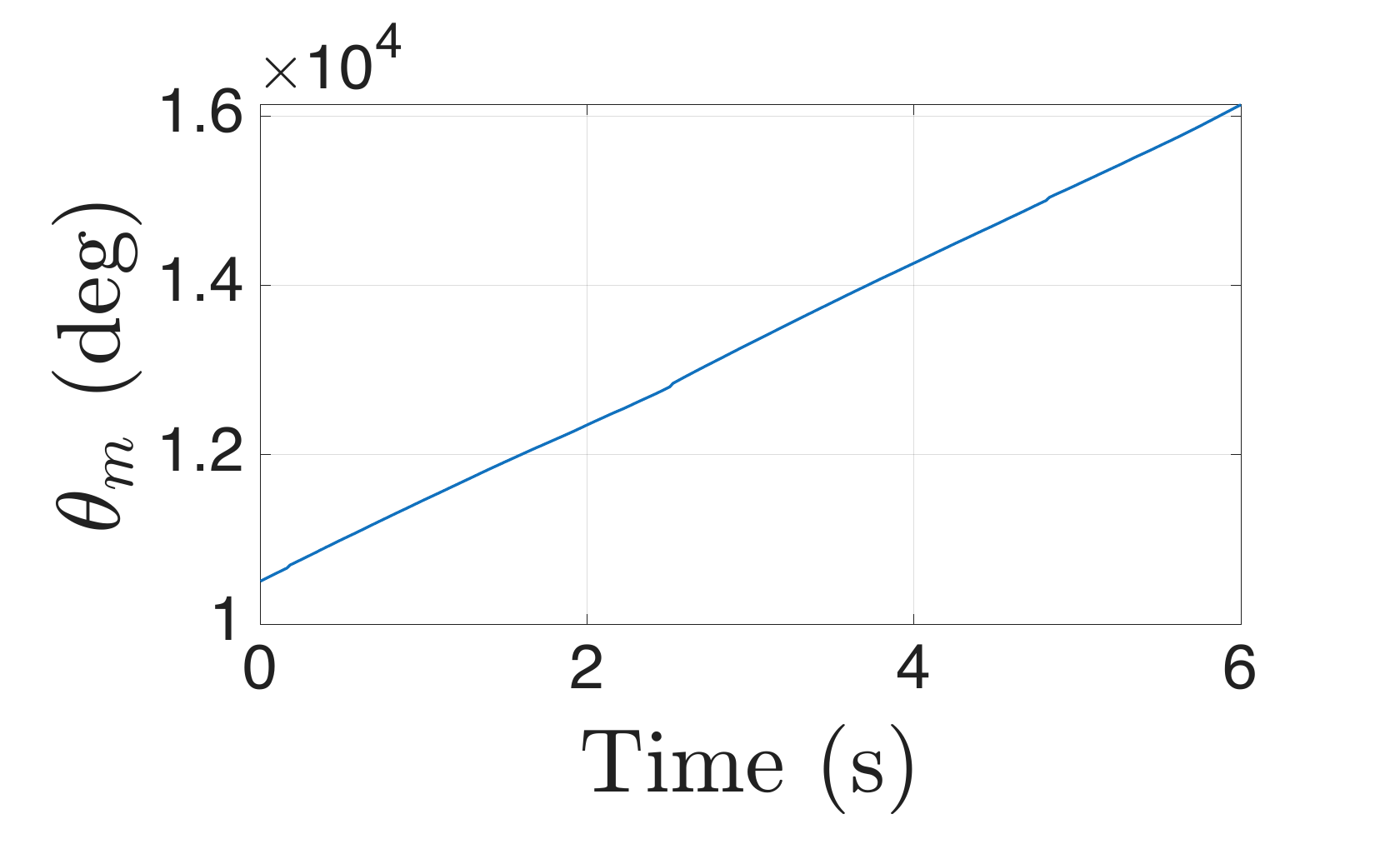}
    \caption{Rotation angle and motor shaft position during rolling movement with origami captured with IMU and motor encoder.}
    \label{fig:IMUdataWithOrigami}
\end{figure}

Fig.~\ref{fig:WithOrigamiOverhead} shows overhead views of the robot lying flat on a surface: (a) without the origami cap, and (b) with the origami cap. In both cases, subfigure (1) corresponds to the neutral, unactuated configuration, where the geometric skeleton is symmetric. As the gearbox rotates, each cable is retracted in sequence, and the four sides contract one at a time in subfigures (2)–(5). This actuation cycle produces an asymmetric deformation of the skeleton and a shape transformation occurs.

The current gearbox design allows each corner to be contracted one at a time. To allow the robot to roll, weights are placed around the structure, so that changing the shape of the skeleton significantly alters the COM based on Eq. (\ref{eq:db_components}). Fig.~\ref{fig:noOrigami4FramesAnnotated} shows the rolling motion where the left side is pulled in by the gearbox, moving the weight on the left towards the center. This shifts the COM to the right and, once the tipping point is reached, the structure rolls to the right. The next pulley is then rotated, causing the left side to contract again and the cycle continues. Fig.~\ref{fig:IMUdataNoOrigami} and Fig.~\ref{fig:IMUdataWithOrigami} show the angle of the robot $\varphi$ and the angle of the motor shaft $\theta_m$ over time for a single rolling motion. Fig.~\ref{fig:noOrigami4FramesAnnotated} corresponds to the motion shown in Fig.~\ref{fig:IMUdataNoOrigami}, while Fig.~\ref{fig:withOrigami4FramesAnnotated} corresponds to the motion shown in Fig.~\ref{fig:IMUdataWithOrigami}. When all corners are contracted simultaneously, a large shape change occurs. By contracting each of the four corners sequentially, the center of mass (COM) of the robot is altered, which induces a rolling motion. The motor rotates, contracting the side until the center of mass has moved far enough to cause the robot to rotate, producing a large change in $\varphi$. Fig.~\ref{fig:IMUdataNoOrigami} (without the origami) shows the structure oscillating after rotating; when the origami is attached the robot has a greater mass which reduces this oscillation. The joints in the origami surface also cause a slower return of the structure to the neutral position, further reducing oscillation. This showcase compliant skeleton alone doesn't have enough damping when do the rotation motion and origami surface works as damper as well to make the motion smoother and controllable. The certain delay between actuation and rotation is caused by the center of mass having to shift past a tipping point, which occurs near the end of the contraction. Fig.~\ref{fig:IMUdataNoOrigami} shows a rotation of \(\varphi\) from \(352^\circ\) to \(82^\circ\) (a complete \(90^\circ\) rotation). With an outer diameter \(D = 188.8\text{mm}\) (radius \(R = 94.4\text{mm}\)), this corresponds to a linear travel distance of \(R \cdot \varphi = 94.4\text{mm} \cdot \tfrac{\pi}{2} \approx 148\text{mm}\). Natually, because the motion release on change on COM this has certain delay betway actuation and rolling motion itself. 

Future iterations of the current gearbox design could add a one-way bearing to each spool. This means that a forward rotation of the driver gear could engage the spools (without the need for actuation of neighboring spools) while a reverse rotation could pick the next spool to be driven, allowing the robot to contract a selected corner based on its current orientation.

\section{Conclusion} 
\label{sec:conclusion}
We presented GeoGami, a mono-actuated origami-inspired platform that combines a geometric compliant skeleton with a compliant origami surface and a cyclic cable drive gearbox (using a single DC motor) to achieve both size compaction (shape transformation) and rolling locomotion from a single input. The transmission uses a worm stage and a sector gear to time-multiplex cable pulls to the four corners; we derived the motor to spool kinematics, mapped cable retraction to side contraction, and formulated a center of mass model that links motor angle to the mass imbalance that drives rolling. We characterized stiffness at the joint and side levels and verified that the longer polygon joints exhibit lower bending stiffness than the shorter ones, while the assembled structure compacts and executes repeatable tipping motions; with 25.1 mm cable retraction per side the prototype achieves approximately \(45^\circ\) rolls and reduced oscillations when the origami cap is installed. Comparative tests of three spindle-based gearboxes (pyramid, 5\,mm, and 10\,mm extrusions) demonstrated that fixed spindle geometries create localized asymmetry and stall once maximum contraction is reached, whereas the cyclic gearbox produces a traveling deformation and sustained progression. Current limitations include tripod locking at extreme contraction, sensitivity to cable stretch and single-layer winding, and incomplete flexibility in some configurations. Future work will refine the transmission with one-way elements and duty cycle tuning, integrate sensing for closed-loop phase control of the traveling pull, and explore materials and crease designs that improve shape recovery and enable continuous locomotion and controllable contact in more complex environments.







 \bibliographystyle{ieeetr}
 \bibliography{references}

\end{document}